\documentclass{article}

\usepackage{graphicx} 
\usepackage{subfigure}
\usepackage{graphics}
\usepackage{epsfig}
\usepackage{multirow}

\usepackage{algorithm}
\usepackage{algorithmic}
\usepackage{appendix}
\usepackage{amsmath}
\usepackage{amssymb}
\usepackage{stmaryrd}
\usepackage[mathscr]{eucal}
\usepackage{lineno}
\usepackage{color}
\usepackage{filecontents}

\usepackage{hyperref}

\def\XBM{\boldsymbol{\mathscr{X}}}
\def\YBM{\boldsymbol{\mathscr{Y}}}

\def\MBM{\boldsymbol{\mathscr{M}}}
\def\ZBM{\boldsymbol{\mathscr{Z}}}

\def\a{{\bf a}}
\def\B{{\bf B}}
\def\C{{\bf C}}

\def\F{{\bf F}}

\def\K{{\bf K}}
\def\H{{\bf H}}

\def\R{{\bf R}}
\def\X{{\bf X}}

\def\S{{\bf S}}

\def\Z{{\bf Z}}
\def\M{{\bf M}}

\def\U{{\bf U}}
\def\u{{\bf u}}

\def\v{{\bf v}}

\def\0{{\bf 0}}
\def\1{{\bf 1}}

\def\RB{{\mathbb R}}

\def\eg{{\emph{e.g. }}}
\def\ie{{\emph{i.e. }}}

\def\argmax{\mathop{\rm argmax}}
\def\argmin{\mathop{\rm argmin}}
\def\tr{\mathsf{tr}}

\def\etal{{\em et al. \/}\,}

\input{macros}

\title{A Report on Multilinear PCA Plus Multilinear LDA to Deal with Tensorial Data: Visual Classification as An Example}

\author{
Shu Kong and Donghui Wang\\
\texttt{\{aimerykong,dhwang\}@zju.edu.cn}\\
College of Computer Science and Technology, Zhejiang University\\
Hangzhou, China
}

\date{\today}

\begin{document}
\maketitle

\begin{abstract}
In practical applications, we often have to deal with high order data,
such as a grayscale image and a video sequence are intrinsically 2nd-order tensor and 3rd-order tensor, respectively.
For doing clustering or classification of these high order data, it is a conventional way to vectorize these data before hand, as PCA or FDA does, which often induce the curse of dimensionality problem.
For this reason, experts have developed many methods to deal with the tensorial data, such as multilinear PCA, multilinear LDA, and so on.
In this paper, we still address the problem of high order data representation and recognition, and propose to study the result of merging multilinear PCA and multilinear LDA into one scenario, we name it \textbf{GDA} for the abbreviation of Generalized Discriminant Analysis.
To evaluate GDA, we perform a series of experiments, and the experimental results demonstrate our GDA outperforms a selection of competing methods such (2D)$^2$PCA, (2D)$^2$LDA, and MDA.
\end{abstract}

\section{Introduction}
\label{sec:intro}
Appearance-based paradigm has been widely employed in the areas of pattern recognition, computer vision and signal processing. One primary advantage of appearance-based methods is that it does not necessarily create representations or models for the objects since, for a given object, its model is implicitly defined by the selection of the samples of the object.
When using appearance-based methods, we usually represent each sample as a vector by vectorizing the data, in other words, we convert an image of size $n\times m$ into a vector of $nm$-dimensional space before hand.
Among various methods dealing with the vector-represented data, principal component analysis (PCA) and linear discriminant analysis (LDA) are the most representative unsupervised and supervised learning method, respectively.

However, when the dimensionality becomes extremely high, PCA and LDA turn to be a time-consuming bottleneck.
Therefore, some researchers and experts seek multidimensional methods that works well in facing with high order data without vectorizing them, such (2D)$^2$PCA~\cite{ZhihuaZhou2D2D} and (2D)$^2$LDA~\cite{Noushath}.

(2D)$^2$PCA and (2D)$^2$LDA have been both shown effective in dealing with 2nd-order data, such as image classification and face recognition.
However, the two methods are limited in 2D data, so when facing higher order data as video data, they will no longer obtain better results. Furthermore, He \etal propose a method to deal with higher-order data, \ie tensor subspace analysis (TSA)~\cite{XiaofeiHeTSA}, which exploits the label information and solves an optimization problem by minimizing the ratio of the within-class scatter and the between-class scatter.
Although they say their algorithm can be easily extended to higher order tensors, it is not convenient to do so due to the optimization formulation they solve.

Tao \etal propose an alternating projection optimization for supervised tensor learning, which is called tensor fisher discriminant analysis, \ie TFDA~\cite{DachengTaoSTL}.
They focuse on the property that when a tensor data is multiplied by a vector, its dimension is reduced by one.
Even though their framework is effective to do high order data recognition when the number of classes is small, it will obtain inferior results when facing large number of classes.
Yan \etal propose multilinear discriminant analysis \ie MDA~\cite{ShuichengYanMDA}, and a novel approach called $k$-mode optimization to iteratively solve the optimization function.
MDA has an advantage in doing high order data recognition, whereas due to relative high dimension in each mode, it is time-consuming in learning process.
Moreover, MDA helps us avoid the curse of dimensionality and alleviate the small sample size problem to some extent,
however, when the training sample size is much smaller than any mode of the sample tensor, their MDA may come across the disaster of singular situation caused by the curse of dimensionality.
Inspired by their framework, we propose to first implement high order SVD (HOSVD) on the tensorial data for dimensionality reduction, and then use multilinear discriminant analysis to learn the most discriminative subspaces of the data.
Actually, HOSVD preserves most valuable information, such as the spatial information and spatial-temporal relationship in video and holds a capability of smoothing, after which the noises in the samples are filtered out to some degree and the dimensions are reduced to a large extent. So the running time in learning process becomes much shorter, and as well, the recognition accuracy will be higher after further dimensionality reduction by multilinear discriminant analysis, as we can see in the experiments. We name the overall process \textbf{GDA} in this paper.

The rest of the paper is organized as follows. Section \ref{sec:Preliminary} explains the notations and gives a brief description of the tensor algebra. Our GDA is described in Section \ref{sec:GFDA}, as well as some details about GDA. Experimental results are presented in Section \ref{sec:ExpResults}. Finally, we conclude our paper in Section \ref{sec:Conclusion}.

\section{Preliminaries}
\label{sec:Preliminary}
Notation for $N$-way arrays can be complex, so we first explain the notations used in this paper, then we review the fundamental algebra of tensors.

\subsection{Notations}
\label{ssec:Notations}
Throughout this paper, ${\bf I}_m$ denotes the $m\times m$ identity matrix.
A high order tensor, a matrix, and a vector are denoted by $\XBM$, ${\bf A}$ and $\a$.
And scalars are denoted by lowercase letters, \eg $a$.

As we frequently use the characters $i$ and $j$ in the meaning of indices, $I$ and $J$ will be reserved to denote the index upper bounds.
The $k^{th}$ elements in a sequence is denoted by a superscript in parentheses, \eg ${\bf A}^{(k)}$, denotes the $n$th matrix in a sequence. When comes to the subscript in parentheses of boldface Euler script letter, such as $\XBM_{(k)}$, one should know it symbolizes a matrix flatten along the $k^{th}$ mode.

 In practice, assume there are $m$ training samples, each sample is represented as the $N$th-order tensors, \ie $\{ \XBM_{i} \in \RB^{I_1 \times I_2 \times \dots \times I_N}, i=1,2,\dots, m \}$, and $\XBM_{i}$ belongs to the class indexed as $c_{i} \in \{1,2,\dots,C \}$ where $C$ is the number of different labels or classes. Consequently, the sample set can be represented as an $(N+1)^{th}$-order sample tensor $\tilde{\XBM} \in \RB^{I_1 \times I_2 \times \dots \times I_N \times m}$.

\subsection{Tensor Algebra}
\label{ssec:TensorAlgebra}
This subsection briefly demonstrates the algebra of tensors which is the fundamental tool in our framework.
A tensor \cite{KoldaTensorDecompApp} is a multi-dimensional array. Flattening a tensor is a kind of converting process of tensor matricization. Specifically, flattening a tensor along the $i$th mode, gives a matrix $\XBM_{(i)}$ in which the columns are resulted from the tensor by varying the value of index $i$, while keeping the other indices fixed.

Tensors can be multiplied together, so obviously the notation and symbols for this are much more complex than matrices. Here we just list some fundamental multiplications related to our work without further introduction and proof \cite{KoldaTensorDecompApp}.
The $k$-$mode$ $(matrix)$ $product$ of a tensor
$\boldsymbol{\mathscr{X}}$
$ \in \RB^{I_{1} \times I_{2} \times \dotsi \times I_{N}}$ with a matrix $\U\in \RB^{J\times I_{k}}$
is denoted by $\boldsymbol{\mathscr{X}}\times {_{k}}{\bf U}$ and is of size
$I_{1} \times \dotsi \times I_{k-1} \times J \times I_{k+1} \times \dotsi \times I_{N}$.
It can be calculated in terms of the flattened form:
\begin{equation}
\boldsymbol{\mathscr{Y}}=\boldsymbol{\mathscr{X}}\times {_{k}}{\bf U}
\Leftrightarrow
\YBM_{(k)} = {\bf U} \XBM_{(k)}.
\label{equa:2}
\end{equation}
Let $\boldsymbol{\mathscr{X}} \in \RB^{I_{1} \times I_{2} \times \dotsi \times I_{N}}$ and ${\bf A}^{(k)}\in \RB^{J_{k} \times I_{k}}$ for all $k\in \{1,\dotsi,N \}$. Then for any $k\in \{1,\dotsi,N \}$, we have
\begin{equation}
\begin{split}
\boldsymbol{\mathscr{Y}}= &
 \boldsymbol{\mathscr{X}}\times {_{1}}{\bf A}^{(1)} \times {_{2}}{\bf A}^{(2)} \times \dotsi \times {_{N}}{\bf A}^{(N)}\\
\Leftrightarrow
\YBM_{(k)}  = &{\bf A}^{(k)}\XBM_{(k)}   {\left( {\bf A}^{(N)} \otimes \dotsi \otimes {\bf A}^{(k+1)} \otimes {\bf A}^{(k-1)} \otimes \dotsi \otimes {\bf A}^{(1)} \right)}^{T}.
\end{split}
\label{equa:3}
\end{equation}
For simplicity, we use
$\llbracket \XBM; {{\bf A}^{(1)}} {{\bf A}^{(2)}},\dots,{{\bf A}^{(N)}} \rrbracket$
to denote Equation \ref{equa:3}, which can be calculated (\ref{equa:3}) in the matrix sense by:
\begin{equation}
\begin{split}
\YBM_{(k)}  = {\bf A}^{(k)}\XBM_{(k)}   {\left( {\bf A}^{(N)} \otimes \dotsi \otimes {\bf A}^{(k+1)} \otimes {\bf A}^{(k-1)} \otimes \dotsi \otimes {\bf A}^{(1)} \right)}^{T}.
\end{split}
\end{equation}
Furthermore, the norm of a tensor is defined as:
\begin{equation}
\Vert\XBM\Vert=\sqrt{\langle\XBM,\XBM\rangle}=
 \Vert\XBM_{(k)}\Vert_{F}
=\sqrt{\sum\limits^{I_1}_{i_1=1} \sum\limits^{I_2}_{i_2=1} \dots \sum\limits^{I_N}_{i_N=1}x^{2}_{i_{1}i_{2}\dots i_{N}}}.
 \end{equation}
So the distance between tensors $\XBM$ and $\YBM$ of the same dimensions is defined as $dist(\XBM,\YBM)=\Vert\XBM-\YBM\Vert$.

\section{Generalized Discriminant Analysis}
\label{sec:GFDA}
In this section, we first review HOSVD. Then we introduce our proposed GDA, followed by \emph{k}-mode optimization.
Finally the full algorithm and classification are presented successively.

\subsection{High Order SVD}
\label{ssec:HOSVD}
In order to generalize SVD for tensors first we have a look at matrix SVD. A matrix $\X$ has two vector spaces: a column space and a row space. SVD decomposes $\X$ into its two vector spaces as: $\X={\bf U}\boldsymbol{\Sigma} {\bf V}^{T}=\boldsymbol{\Sigma} \times _{1}{\bf U} \times _{2}{\bf V}$, where ${\bf U}$ and ${\bf V}$ represent the orthogonal column space and row space respectively. Then we come to SVD for tensors which have $N$ associated vector spaces.
HOSVD decomposes the tensor $\boldsymbol{\mathscr{X}}\in \RB^{I_{1} \times \dots \times I_{N}}$ into its $N$ vector spaces by:
\begin{equation}
\boldsymbol{\mathscr{X}} \approx \boldsymbol{\mathscr{Y}}\times {_{1}}{\bf V}^{(1)}  \times {_{2}}{\bf V}^{(2)} \times \dotsi \times {_{N}}{\bf V}^{(N)},
\label{equa:4}
\end{equation}
where ${\bf V}^{(k)} \in {\RB}^{I_{k}\times J_{k}}$,
${\bf V}^{(k)^T} {\bf V}^{(k)} = {\bf I}_{J_{k}}$
and $\YBM \in \RB^{J_{1} \times \dots \times J_{N}}$.
Here ${\bf V}^{(k)}$ in which $k\in\{1,2,\dotsi,N \}$ represents the $k$-mode vector spaces,
and $\boldsymbol{\mathscr{Y}}$ is the core tensor which shows the interaction between different spaces.

According to Equation (\ref{equa:3}) and (\ref{equa:4}), the $N$-mode SVD algorithm \cite{KoldaTensorDecompApp,Vasilescu2002} for decomposing tensor $\boldsymbol{\mathscr{X}}$ is:

1) For $k=1,\dotsi,N$ compute the SVD of $\XBM_{(k)}$, ${\bf V}^{(k)} \in \RB^{ I_{k} \times J_{k}}$ is the left singular vectors of $\XBM_{(k)}$, where $J_{k}$ can be chosen less than $I_{k}$ by a kind of criterion;

2) The core tensor $\boldsymbol{\mathscr{Y}}$ is computed by:
\begin{equation}
\boldsymbol{\mathscr{Y}} \approx \boldsymbol{\mathscr{X}}\times {_{1}} {\bf V}^{(1)^{T}}  \times _{2}{\bf V}^{(2)^{T}} \times \dotsi \times _{N}{\bf V}^{(N)^{T}}.
\label{equa:5}
\end{equation}

\subsection{Multilinear Discriminant Analysis and \emph{k}-Mode Optimization}
\label{ssec:kModeOpt}
In order to find the new tensor space that maximize the ratio of the between-class scatter and the within-class scatter, we tend to solve the optimization function:
\begin{equation}
\begin{split}
&{\bf U}^{(k)}|^{N}_{k=1} = \argmax\limits_{{\bf U}^{(k)}|^{N}_{k=1}} \frac{\sum\limits^{C}_{i=1}n_{i}\Vert \llbracket\MBM_{i}-\tilde{\MBM}; {\bf U}^{(1)^{T}},{\bf U}^{(2)^{T}},\dots,{\bf U}^{(N)^{T}}\rrbracket\Vert^2}
{\sum\limits^{C}_{i=1}  \sum\limits_{j\in C_{i}}\Vert\llbracket \XBM_{j} - \MBM_{i}; {\bf U}^{(1)^{T}},{\bf U}^{(2)^{T}},\dots,{\bf U}^{(N)^{T}}\rrbracket\Vert^2},
\end{split}
\label{equa:OptFunction}
\end{equation}
where $\MBM_{i}$ and $\tilde{\MBM}$ represent the mean of the $i^{th}$ class and the global mean of the training data, respectively. However, the objective function (\ref{equa:OptFunction}) has no closed-form solution due to that the ${\bf U}^{(k)}|^{N}_{k=1}$ depends on each other, so we have to solve (\ref{equa:OptFunction}) by an iterative procedure. Having noticed $\Vert\XBM\Vert = \Vert\XBM_{(k)}\Vert_{F}$ and $\Vert\X\Vert^{2}=Tr(\X^{T}\X)=Tr(\X\X^{T})$, and if we assume ${\bf U}^{(1)},\dots,{\bf U}^{(k-1)},{\bf U}^{(k+1)},\dots,{\bf U}^{(N)}$ are already known, then we can calculate ${\bf U}^{(k)}$ by:
\begin{equation}
\begin{split}
{\bf U}^{(k)}
& = \argmax_{{\bf U}_{k}}
{\frac
{\tr \left( {\bf U}^{(k)^{T}} {\S}_{B(k)} {\bf U}^{(k)} \right)}
{\tr \left( {\bf U}^{(k)^{T}} {\S}_{W(k)} {\bf U}^{(k)} \right)}}.
\end{split}
\label{equa:kModeOpt}
\end{equation}
Denote ${\bf U}_{p}={\bf U}^{(N)^{T}} \otimes \dots \otimes {\bf U}^{(k+1)^{T}} \otimes {\bf U}^{(k-1)^{T}} \otimes \dots \otimes {\bf U}^{(1)^{T}}$, then ${\S}_{B(k)}$ and ${\S}_{W(k)}$  are between-class and the within-class scatter matrix along the $k^{th}$ mode respectively:
\begin{align}
\begin{split}
{\S}_{B(k)} &= \sum\limits^{C}_{i=1} n_{i}\left((\MBM_{i}-\tilde{\MBM})_{(k)}\right)
{\bf U}_{p}^{T}{\bf U}_{p} \left((\MBM_{i}-\tilde{\MBM})_{(k)}\right)^{T},\\
\S_{W(k)} &= \sum\limits^{C}_{i=1}  \sum\limits_{j\in C_{i}}
\left((\XBM_{j}-\MBM_{i})_{(k)}\right)
{\bf U}_{p}^{T}{\bf U}_{p} \left((\XBM_{j}-\MBM_{i})_{(k)}\right)^{T}.
\nonumber
\end{split}
\end{align}
The optimal projected subspace along mode-$k$ is spanned by the columns of ${\bf U}^{(k)}$, which is the solution of equation (\ref{equa:kModeOpt}).
It is easy to explicitly solve the singular problem of $\S_{W(k)}^{-1}\S_{B(k)}$ to obtain ${\bf U}^{(k)}$.
The iterative procedure to solved (\ref{equa:OptFunction}) is called $k$-mode optimization, which is first put forward by Yan \etal \cite{ShuichengYanMDA}. We continue to use this term for its conciseness.

\subsection{Generalized Discriminant Analysis --- Algorithmic Analysis}
\label{ssec:AlgorithmicAnalysis}
If we directly iteratively solve ${\bf U}^{(k)}$, the learning process will be time-consuming and we may meet the curse of dimensionality. Like Fisherface \cite{PeterEigenfaceFisherface}, we first do the dimensionality reduction of the data set via HOSVD. In this stage, we choose a threshold $\theta$ to achieve dimensionality reduction purpose:
\begin{equation}
\frac{\sum\limits_{i=1}^{d}{\sigma_{i}}}{ \sum\limits_{i=1}^{M}{\sigma_{i}}} \ge \theta
\label{equa:ThresholdDimReduction}
\end{equation}
 where $\sigma_{1}$,$\dotsi$,$\sigma_{d}$ is the $d$ largest singular values of $\XBM_{(k)}$. However HOSVD does not exactly mean deleting the rest $1-\theta$ features, instead it smoothes the samples by filtering out random noises to some extent, as Fig.\ref{fig:RestoreWeizmannWalk} shows the smoothing result from the silhouettes of a video clip.

After HOSVD, we come to solve Equation (\ref{equa:OptFunction}).
The lower dimension enables us to speed up the training process and avoid the singular situation.
The whole algorithm are displayed Algorithm \ref{alg:GFDA}, in which we use the superscript $t$ to denote the
resulting ${\bf U}^{(k)}$ of the $t^{th}$ iteration.

\begin{algorithm}[!t]
\caption{Generalized Fisher Discriminant Analysis}
\label{alg:GFDA}
\emph{Input}: The training set $\tilde{\XBM} \in \RB^{I_{1} \times I_{2} \times \dots\times I_{N} \times m}$, their class labels $c_{i} \in \{ 1,2,\dots,C \}$, where $i =1,2,\dots,m$, and the final lower dimensions $I_{1}^{'} \times I_{2}^{'} \times \dots\times I_{N}^{'}$.

\emph{Output}: the projectors $\left({\bf V}^{(k)}{\bf U}^{(k)}\right) \in \RB^{I_{k} \times I_{k}^{'}}$, where $I_{k}^{'} < I_{k}$ and $k\in\{1,2,\dots,N\}$.
\begin{algorithmic}[1]

\STATE  Use high order SVD to decompose the training set $\tilde{\XBM}$ as:
        \begin{equation}
        \tilde{\XBM}\approx\tilde{\YBM} \times_{1}{\bf V}^{(1)}\times\dots\times_{N}{\bf V}^{(N)}\times_{N+1}{\bf I}_m
        \nonumber
        \end{equation}
        where $\tilde{\YBM} \in \RB^{J_{1} \times J_{2} \times \dotsi \times J_{N} \times m}$ and ${\bf V}^{(k)}\in\RB^{I_{k}\times J_{k}}$

\STATE  Initialize ${\bf U}^{(k)}\in \RB^{J_{k} \times I_{k}^{'}}$, where $k\in\{1,2,\dots,N\}$;

\STATE  Calculate the projected mean of each class $\MBM_{i}$ and the projected global mean $\tilde{\MBM}$;

\WHILE{ stop criterion is not reached }
    \FOR{$k=1,2,\dots,N$}
\STATE
\begin{displaymath}
\begin{split}
{\bf U}_{p}=&{\bf U}^{(N),{t}^{T}} \otimes \dots \otimes {\bf U}^{(k+1),{t}^{T}} \otimes {\bf U}^{(k-1),{t+1}^{T}} \otimes
        \dots  \otimes {\bf U}^{(1),{t+1}^{T}}
\end{split}
\end{displaymath}
\begin{displaymath}
\begin{split}
\S_{B(k)}  = &\sum\limits^{C}_{i=1} n_{i}
 \left((\MBM_{i}-\tilde{\MBM})_{(k)}\right)
{\bf U}_{p}^{T}{\bf U}_{p} \left((\MBM_{i}-\tilde{\MBM})_{(k)}\right)^{T}
\end{split}
\end{displaymath}
\begin{displaymath}
    \begin{split}
         \S_{W(k)} = &\sum\limits^{C}_{i=1}\sum\limits_{j\in C_{i}}
                 \left((\YBM_{j}-\MBM_{i})_{(k)}\right)
                    {\bf U}_{p}^{T}{\bf U}_{p} \left((\YBM_{j}-\MBM_{i})_{(k)}\right)^{T}
          \nonumber
          \end{split}
    \end{displaymath}
         \STATE Solve the optimization problem through generalized eigenproblem:
         \begin{displaymath}
                    {\bf U}^{(k),{t+1}}= \arg\max_{{\bf U}^{(k)}}
                    {\frac
                    {\tr \left( {\bf U}^{(k),{t+1}^{T}} \S_{B(k)} {\bf U}^{(k),{t+1}} \right)}
                    {\tr \left( {\bf U}^{(k),{t+1}^{T}} \S_{W(k)} {\bf U}^{(k),{t+1}} \right)}}
         \end{displaymath}
         ${\bf U}^{(k),{t+1}}$ is the $I^{'}_{k}$ top eigenvectors of $\S^{-1}_{W(k)}\S_{B(k)}$.
    \ENDFOR
\ENDWHILE
\end{algorithmic}
\end{algorithm}

\subsection{Classification With GDA}
\label{ssec:ClassificationWithGFDA}
With the learned projectors $\{{\bf V}^{(k)}|_{k=1}^{N}\}$ and $\{{\bf U}^{(k)}|_{k=1}^{N}\}$, the low-dimensional representation of the training sample $\XBM_{i}$, $i=1,2,\dots,m$, can be computed as $\ZBM_{i}=\XBM_{i} \times _{1}\left({\bf V}^{(1)}{\bf U}^{(1)}\right)^{T}\times \dots \times _{N}\left({\bf V}^{(N)}{\bf U}^{(N)}\right)^{T}$. When a new data $\XBM$ comes, we first compute its low-dimensional representation as
$\ZBM=\llbracket \XBM; \left({\bf V}^{(1)}{\bf U}^{(1)}\right)^{T},  \dots, \left({\bf V}^{(N)}{\bf U}^{(N)}\right)^{T} \rrbracket$.
Then its class label is predicted to be that of the sample whose low-dimensional representation is nearest to $\ZBM$, that is $c_{i^{*}}$ where
\begin{equation}
i^{*}=\argmin\limits_{i}\Vert \ZBM-\ZBM_{i} \Vert
\end{equation}
 In this paper, we use this \emph{nearest-neighbor} method for the final classification throughout all the experiments owing to its simplicity in computation.

\section{Experimental Results}
\label{sec:ExpResults}
In this section, we conduct a series of experiments to consider the performance of our proposed GDA in dimensionality reduction, clustering and recognition. All of our experiments are carried out on a PC machine with Pentium(R) Dual-Core CPU and 4.00G memory.
\subsection{Data Preparation}
\label{ssec:DataPreparation}
\begin{figure}[t]
\begin{center}
\includegraphics[width=0.9\linewidth]{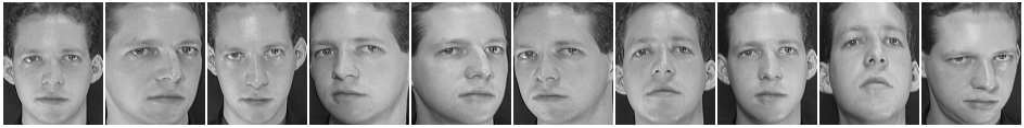}
\end{center}
   \caption{10 samples in the ORL face database.}
\label{fig:ORLsamples}
\end{figure}

Two benchmark databases, ORL \cite{ORL} and Weizmann \cite{ActionsAsSpaceTimeShapes_iccv05} are used in our experiments.
ORL database contains 400 images of 40 individuals and each image is grayscale
and normalized to the resolution of 112*92 pixels.
In ORL database, the images are taken at different times, varying the lighting, facial expressions (open or closed eyes, smiling or not smiling) and facial details (glasses or no glasses).
All the images are taken against a dark homogeneous background with the subjects in an upright, frontal position (with tolerance for some side movement).
Fig.\ref{fig:ORLsamples} illustrates 10 images of one individual from ORL database.

Weizmann database is a recent database with a reasonable size reported in \cite{ActionsAsSpaceTimeShapes_iccv05}.
It contains ten action classes performed by nine individuals.
The actions include bending (bend), jumping jack (jack), jumping-forward-on-two-legs (jump), jumping-in-place-on-two-legs (pjump), running (run), galloping sideways (side), skipping (skip), walking (walk), waving-one-hand (wave1), and waving-two-hands (wave2).
Hence, we have 90 video sequences in all.
For action recognition experiment, we directly use the binary silhouettes of Weizmann database. Some of these silhouettes are deformed and noisy due to segmentation problems, but be still contained in the training set. In Weizmann database, each action video generally includes $2$-$4$ complete action cycles. Using a single period is much more computationally efficient than using the entire length of the video. Also we need to have equal-length sequences in our tensor framework. So we find the period for all the training action sequences such as $10$ frames, which may be smaller than maximum period, but this problem can be solved by deleting the extra frames randomly.

\subsection{Dimensionality Reduction and Smoothing}
\label{ssec:Smoothing}

In this subsection, we conduct an experiment to show HOPCA (HOSVD), which is a part of our proposed GDA, can do better in dimensionality reduction, and preserve spatial information of images and spatiotemporal relationship of videos. Furthermore, HOPCA can filter out the random noises in some sense.

Essentially, (2D)$^2$PCA \cite{ZhihuaZhou2D2D} is 2nd-order PCA, so in this case, HOPCA equals to (2D)$^2$PCA.
In this experiment, we need to have a look at CR and PSNR, which denote compression ratio and peak signal-to-noise ratio, respectively.
CR is defined as the ratio of the compressed size to the uncompressed size, and the formula for calculating PSNR is:
\begin{equation}
PSNR=20\log_{10} (\frac{255}{\sqrt{\Vert \X_{noised}-\X_{original} \Vert_{F}^{2}/mn}})
\nonumber
\end{equation}
where $\X_{noised}$ and $\X_{original}$ are both matrices of the size $m\times n$.
Suppose there are $M$ training face images with size $m \times n$, the number of projection vectors in PCA and HOPCA is p, d and q. Then the compression ratios of PCA and HOPCA are computed as $Mmn/(Mp+mnp)$ and $Mmn/(Mdq+md+nq)$ respectively.

\begin{figure}[htb]
\centering
    \includegraphics[width=0.9\linewidth]{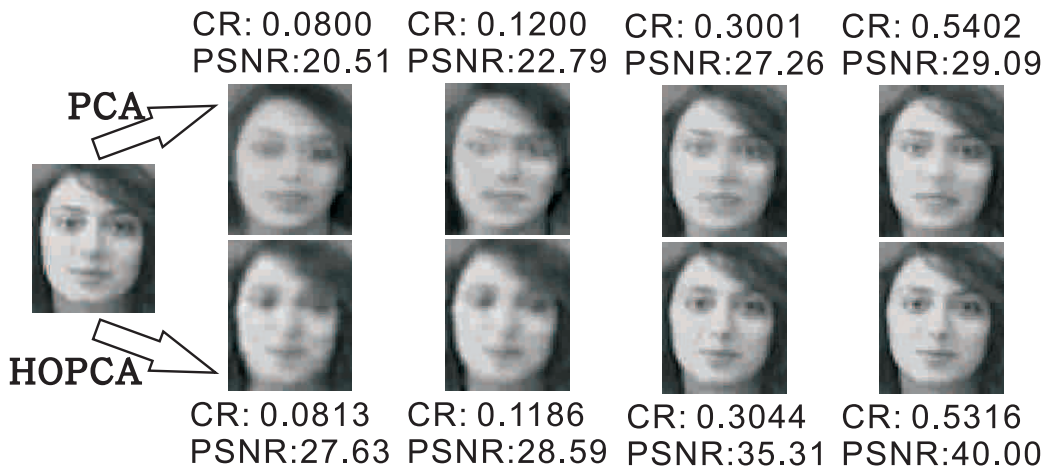}
  \medskip
\caption{ Reconstructed image from ORL database by PCA and HOPCA.
The original image is shown on the leftmost. The first row shows a series image of restoration from PCA reduction, and the second row displays that from HOPCA reduction.
}
\label{fig:RestorationORLFace}
\end{figure}

We restore the images from the dimensionality reduced dataset. Fig.~\ref{fig:RestorationORLFace} shows some reconstruction results under similar compression ratios. It is obvious that HOPCA preserves more inherent characteristics.

\begin{figure}[htb]
\centering
    \begin{minipage}[t]{0.7\linewidth}
        \centering
        \includegraphics[width=1\linewidth]{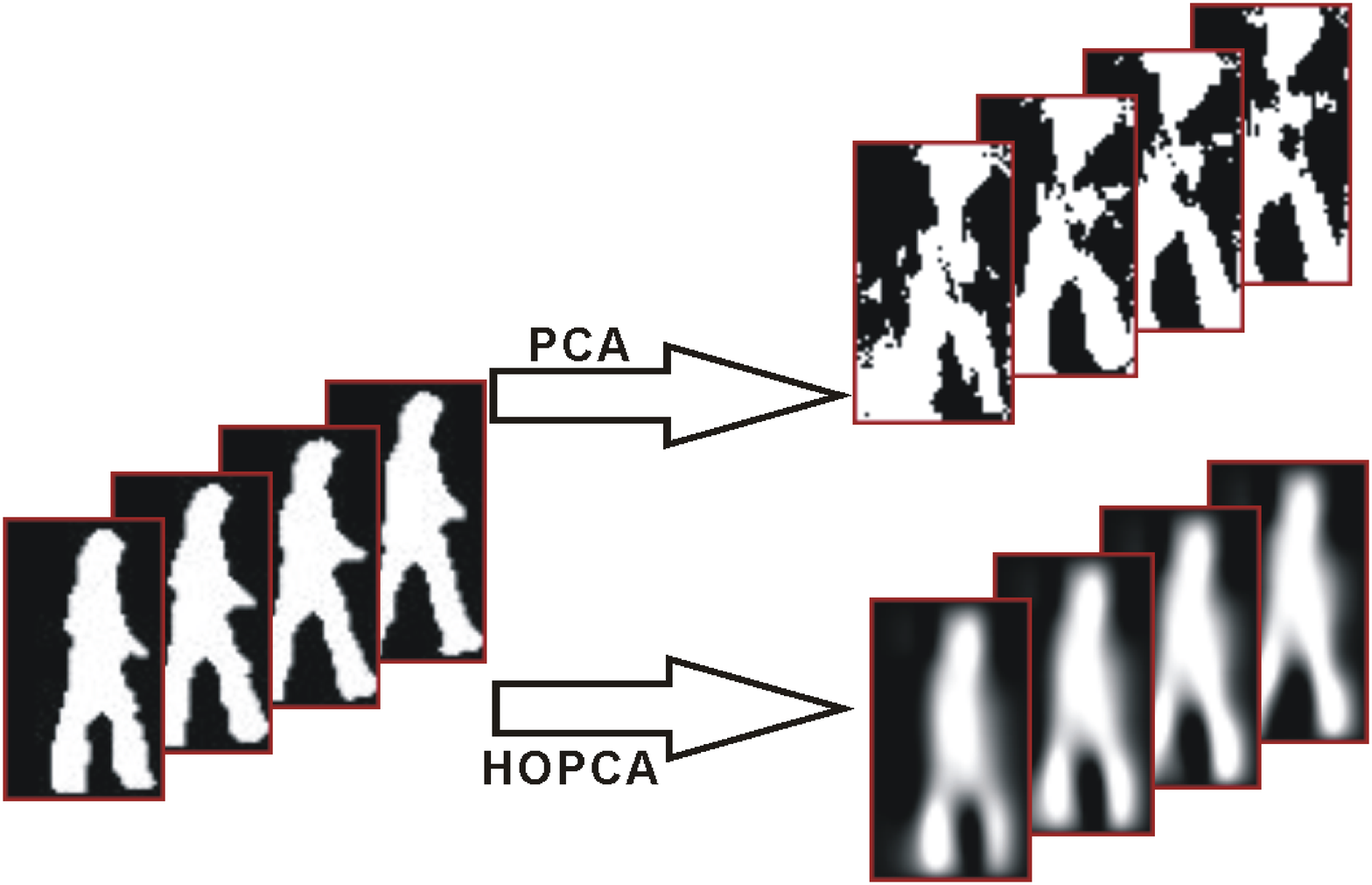}
    \end{minipage}
\caption{One reconstructed training video sequence on Weizmann database under similar compression ratio by PCA and HOPCA.
Here $\theta$ is set as 0.95.
It is obvious that PCA destroys the spatial-temporal information in the video, causing severe distortion, whereas HOPCA preserves the space-time relationship.
}
\label{fig:RestoreWeizmannWalk}
\end{figure}

Furthermore, we conduct experiments to see the abilities of PCA and HOPCA in representing action videos under similar compression ratios. We randomly select a period of video sequence in Weizmann database, and for simplicity, we use the binary silhouette of the sequence here.
The comparison shows HOPCA preserves spatial-temporal relationships when dealing with video compression, whereas PCA destroys the important information, as Fig.~\ref{fig:RestoreWeizmannWalk} shows.

From this experiment, we can easily see HOSVD not only accomplishes dimensionality reduction better than that of PCA, but also preserves more valuable intrinsic structures such spatial and spatial-temporal relationships, which PCA will destroy. Therefore high order SVD can filter the noises out in some sense, which will boost the computation and enhance the performance in the recognition process.

\subsection{Visualization of Dataset --- Clustering and Classification}
\label{ssec:Visualization}

In this subsection, we use PCA, (2D)$^2$PCA, MDA and GDA to project the images into a 2-dimensional subspace for visualization. This experiment helps us understand that our proposed GDA can obtain more discriminating power than the other 3 methods under a relative lower dimensional subspace.

\begin{figure*}
\centering
\begin{minipage}[b]{.3\linewidth}
  \centering
  \centerline{\epsfig{figure=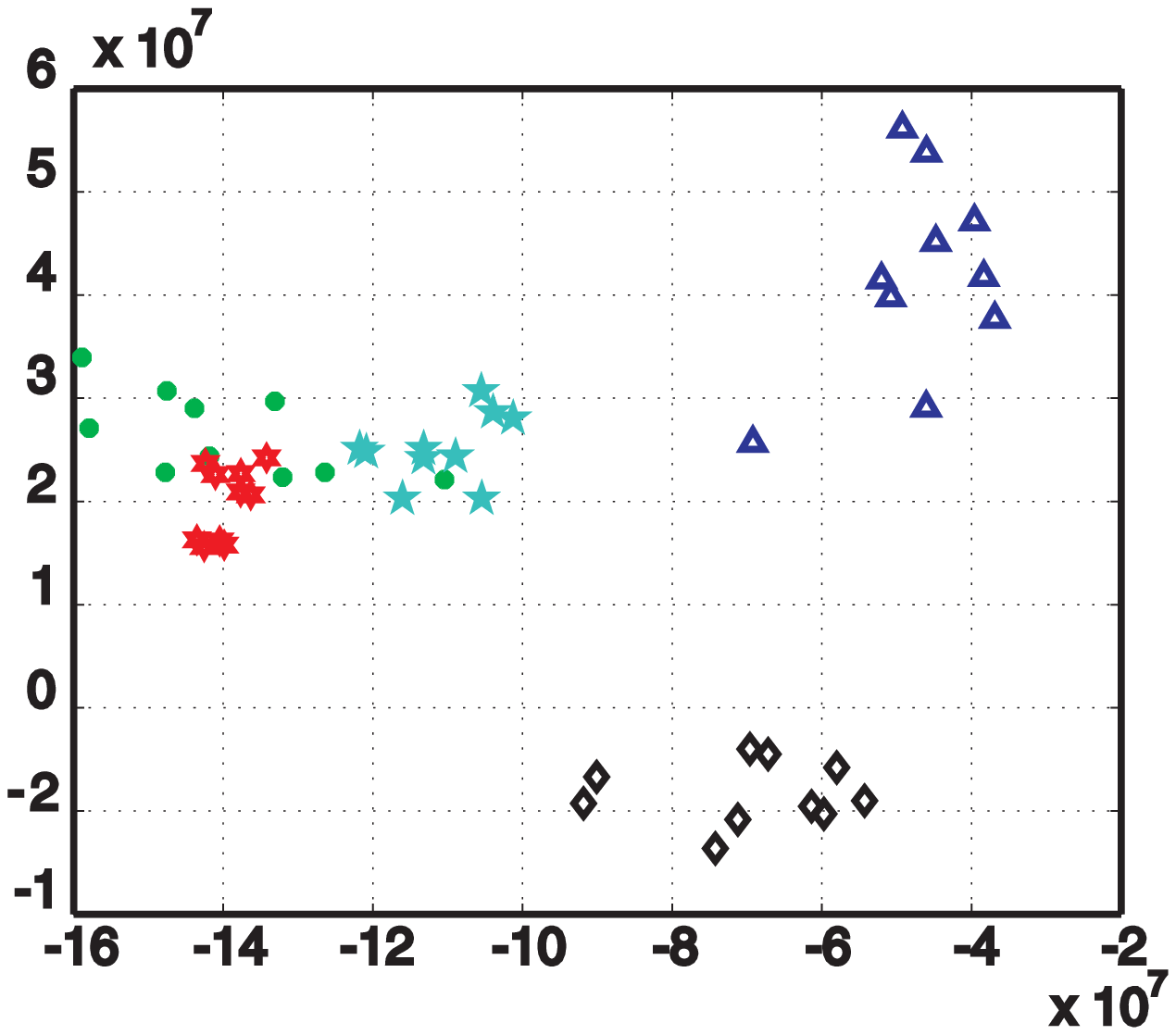,width=5cm}}
  \vspace{0.0cm}
  \centerline{(a) PCA on $\RB^{2}$}\medskip
\end{minipage}
\begin{minipage}[b]{.3\linewidth}
  \centering
  \centerline{\epsfig{figure=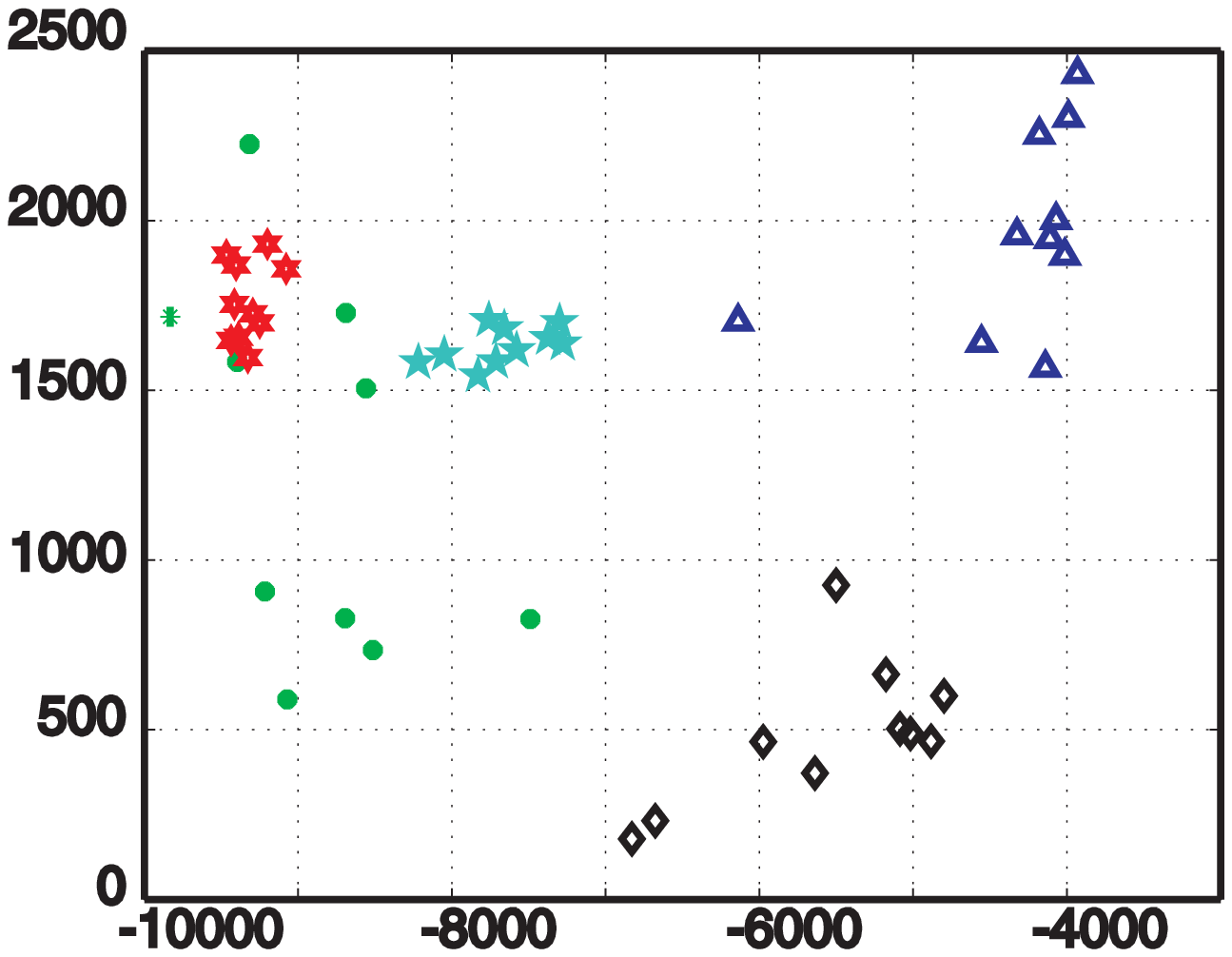,width=5cm}}
  \vspace{0.0cm}
  \centerline{(b) (2D)$^2$PCA on $\RB^{1\times2}$}\medskip
\end{minipage}
\begin{minipage}[b]{0.3\linewidth}
  \centering
  \centerline{\epsfig{figure=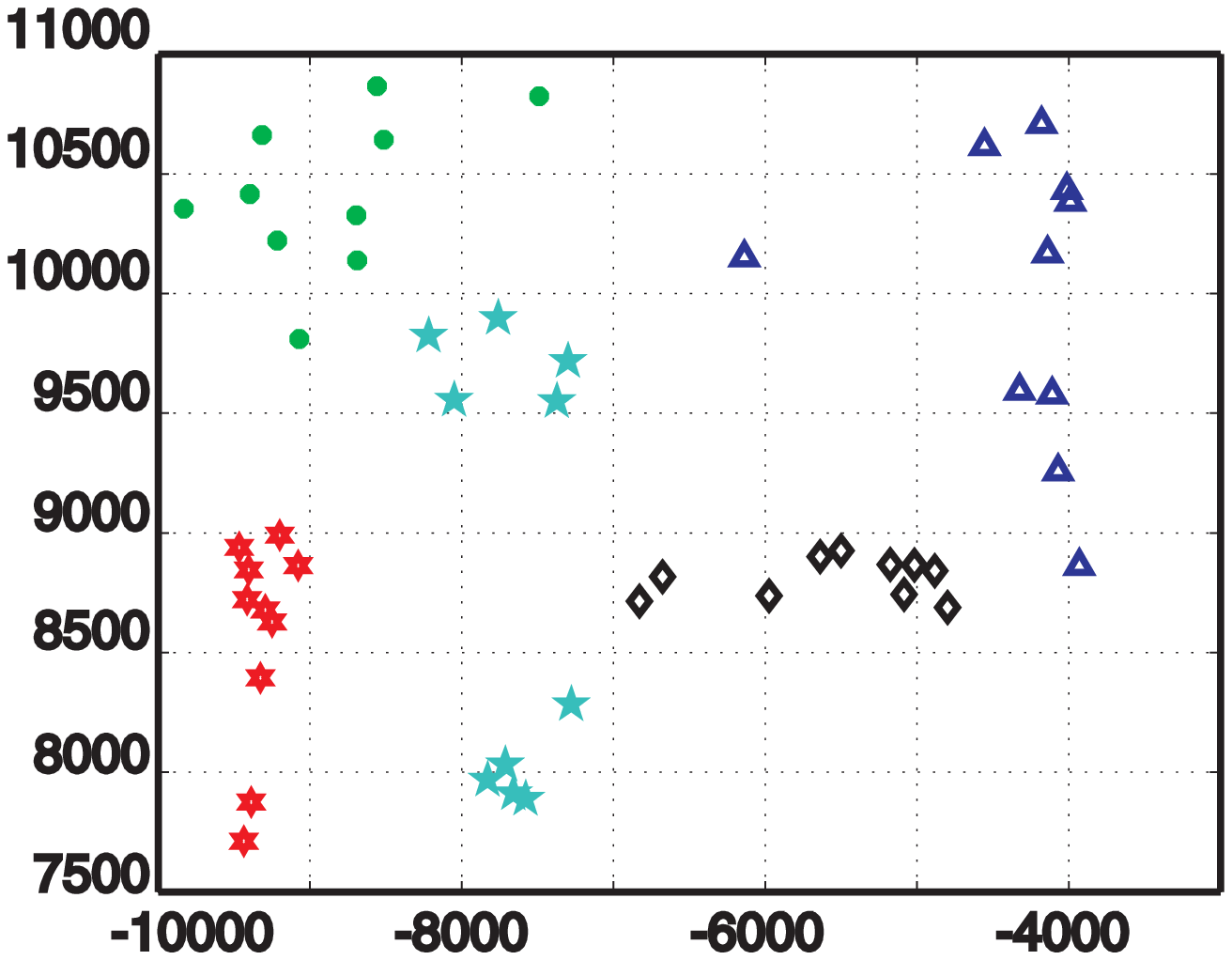,width=5cm}}
  \vspace{0.0cm}
  \centerline{(c) (2D)$^2$PCA on $\RB^{2\times1}$}\medskip
\end{minipage}
\begin{minipage}[b]{.47\linewidth}
  \centering
  \centerline{\epsfig{figure=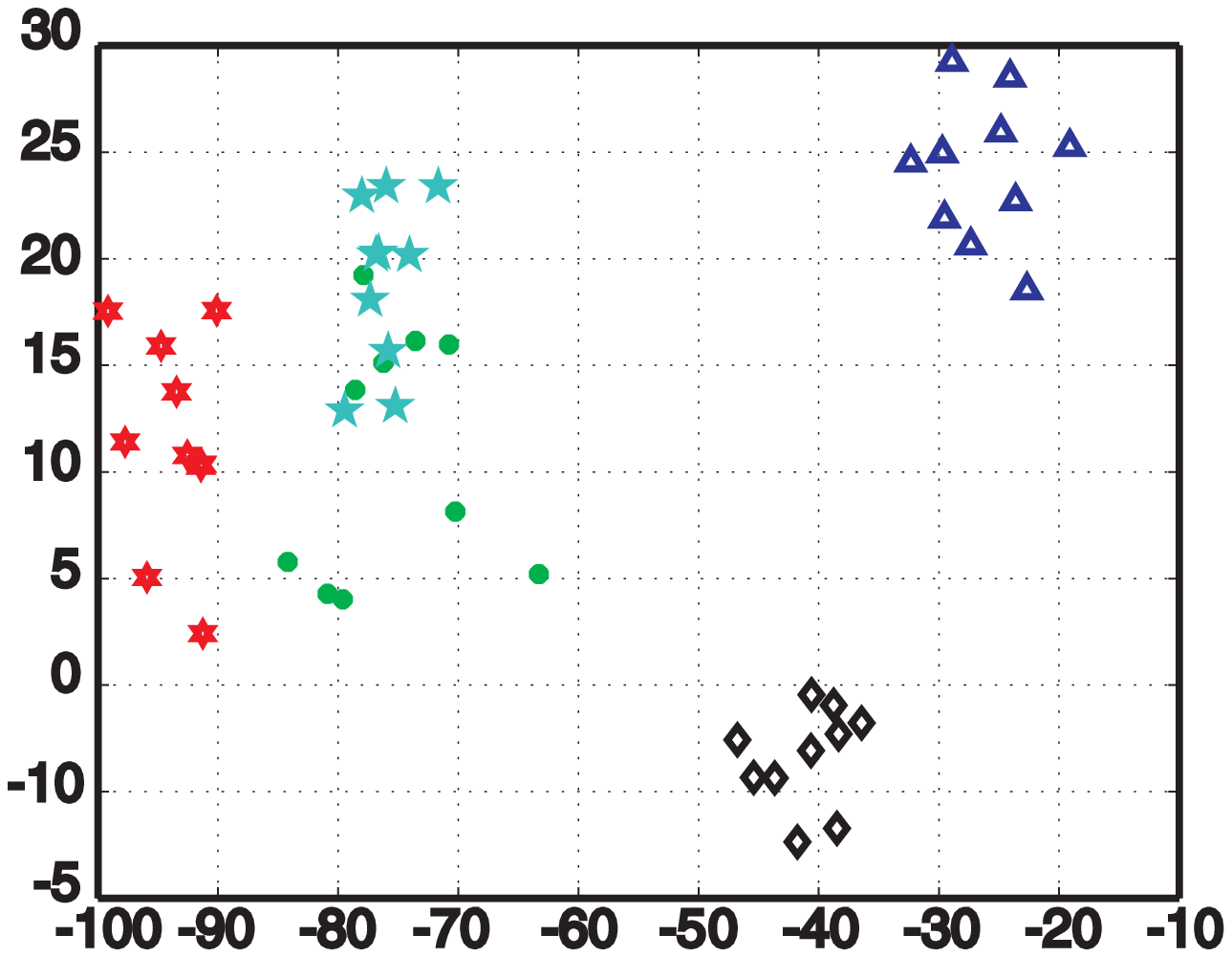,width=7cm}}
  \vspace{0.0cm}
  \centerline{(d) MDA on $\RB^{1\times2}$}\medskip
\end{minipage}
\begin{minipage}[b]{0.47\linewidth}
  \centering
  \centerline{\epsfig{figure=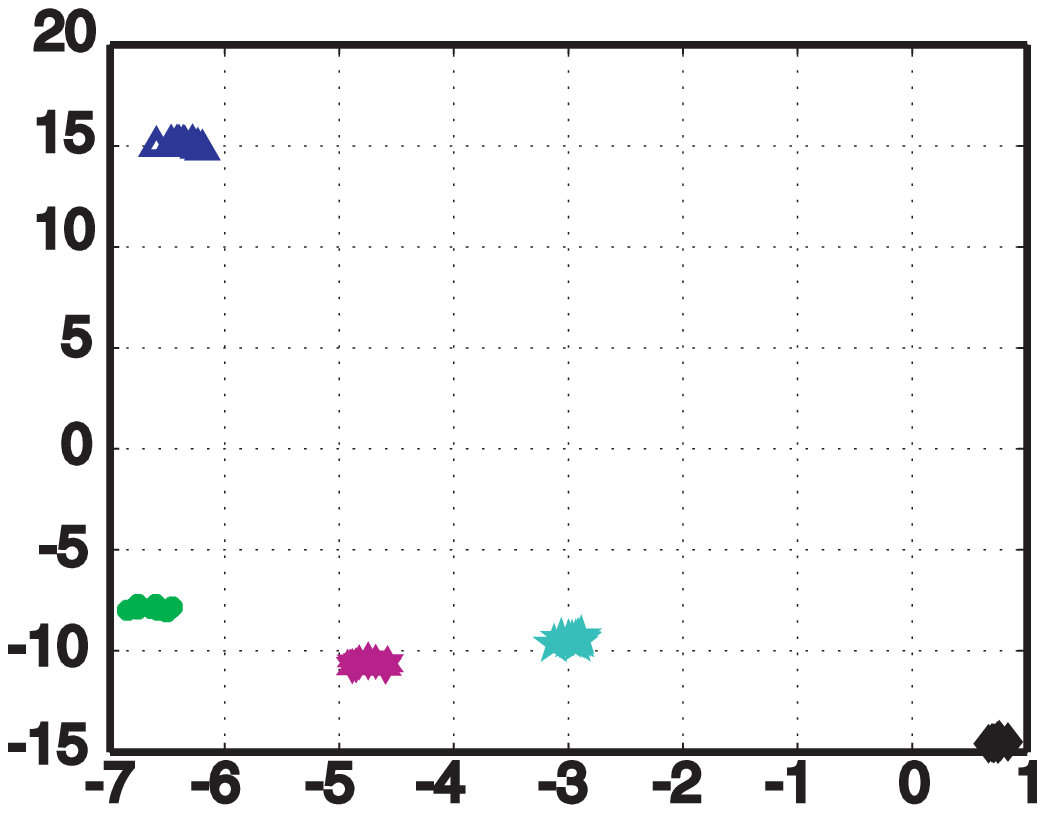,width=7cm}}
  \vspace{0.0cm}
  \centerline{(e) MDA on $\RB^{2\times1}$}\medskip
\end{minipage}
\begin{minipage}[b]{.47\linewidth}
  \centering
  \centerline{\epsfig{figure=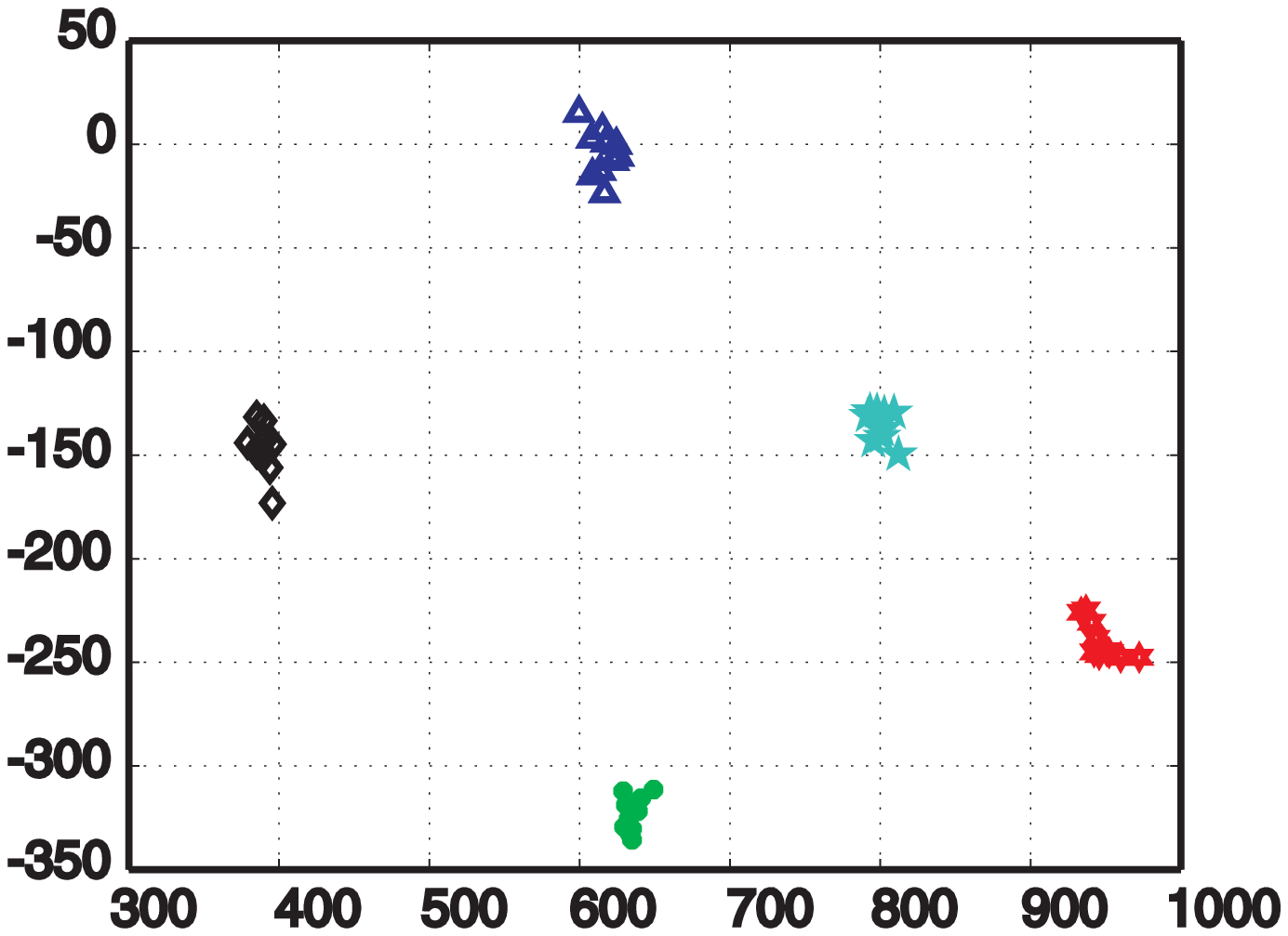,width=7cm}}
  \vspace{0.0cm}
  \centerline{(f) GDA on $\RB^{1\times2}$}\medskip
\end{minipage}
\begin{minipage}[b]{0.47\linewidth}
  \centering
  \centerline{\epsfig{figure=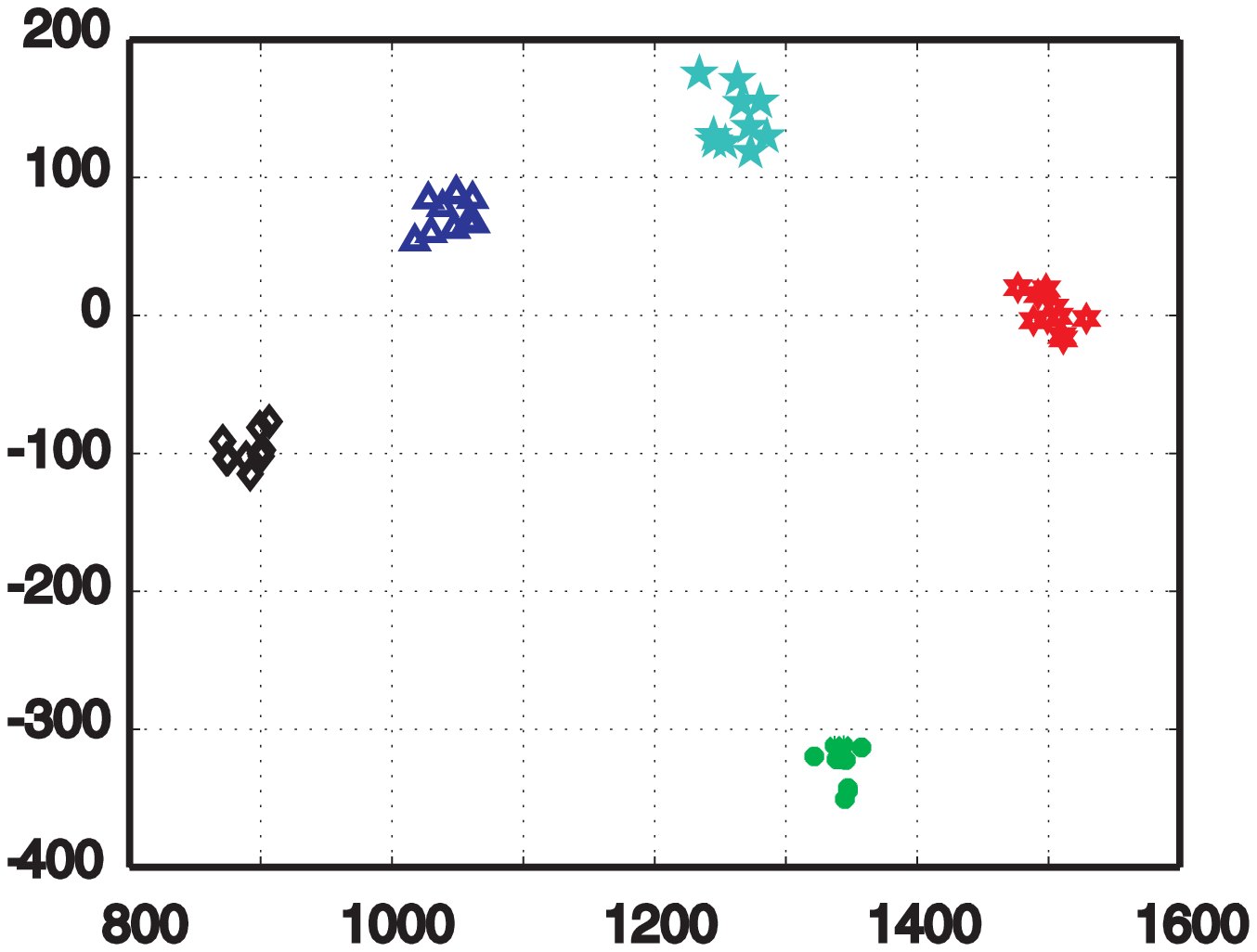,width=7cm}}
  \vspace{0.0cm}
  \centerline{(g) GDA on $\RB^{2\times1}$}\medskip
\end{minipage}
%
\caption{ 2D visualization of five classes in ORL dataset.
The HOSVD threshold (Equation.\ref{equa:ThresholdDimReduction}) in GDA is set 0.994
}
\label{fig:Projection}
\end{figure*}

We select the first 5 individuals in ORL for this test and Fig.~\ref{fig:Projection} shows the results. For PCA, we select the two eigenvectors corresponding to the first two largest eigenvalues of the covariance matrix, projecting the images into 2-dimensional subspace. And for (2D)$^2$PCA, MDA and GDA, we project the images into either $\RB^{1\times2}$ and
$\RB^{2 \times 1}$, both of which are 2-dimensional spaces.
$\RB^{1\times2}$ is formed by the projection $\u^T_1 \X [\v_1, \v_2]$ and $\RB^{2\times1}$
is formed by the projection $[\u_1,\u_2]^T\X \v_1$, here $\X$ stands for a image, and $\u$ and $\v$ are the columns of projectors.

As can be seen from Fig.~\ref{fig:Projection}, PCA performs the worst, it fails to distinguish the different classes from a clustering viewpoint. In contrary, (2D)$^2$PCA clusters the data better than PCA, even if it also mixes some classes together.
MDA, as a supervised method, does significantly well than PCA and (2D)$^2$PCA, which can be seen from (d) and (e). As well, MDA also mixes some categories together, seen in Fig.~\ref{fig:Projection} (d).
However, one fatal drawback of MDA is the between-class distance is much smaller than that of GDA, it may lead to inferior recognition, in the sense of overfitting.
Clearly our GDA performs the best, it maintains a good distance between each classes, and also keeps the a good within-class aggregation.
This illustrative example shows that GDA can have more discriminating power than others under a relatively lower dimension.

\subsection{Classification Result on ORL Database}
\label{ssec:ORLExp}

This subsection shows a set of experiments to test the performance of our GDA in face recognition, in which each image is represented as a 2nd-order tensor.

Three sets of experiments were conducted to compare the face recognition performance of GDA with PCA, (2D)$^2$PCA, (2D)$^2$LDA and MDA. For ease of representation, the experiments are named as Train$m$/Test$n$ which means that $m$ images of per person are \emph{randomly} selected for training and the remaining $n$ images for testing.

\begin{figure*}
\centering
\begin{minipage}[b]{.48\linewidth}
  \centering
  \centerline{\epsfig{figure=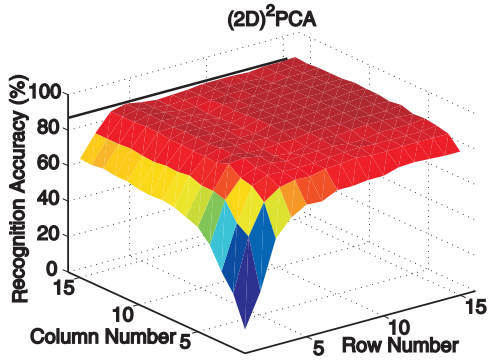,width=7cm}}
  \vspace{0.0cm}
  \centerline{(a) (2D)$^2$PCA original view}\medskip
\end{minipage}
\hfill
\begin{minipage}[b]{0.48\linewidth}
  \centering
  \centerline{\epsfig{figure=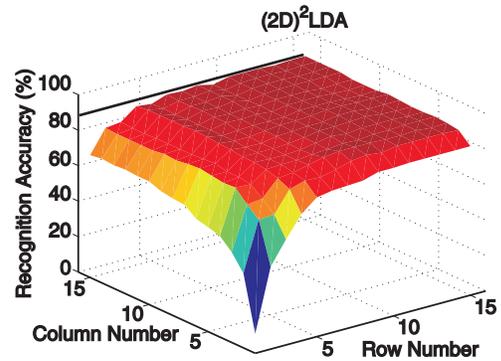,width=7cm}}
  \vspace{0.0cm}
  \centerline{(c) (2D)$^2$LDA original view}\medskip
\end{minipage}
\hfill
\begin{minipage}[b]{0.48\linewidth}
  \centering
  \centerline{\epsfig{figure=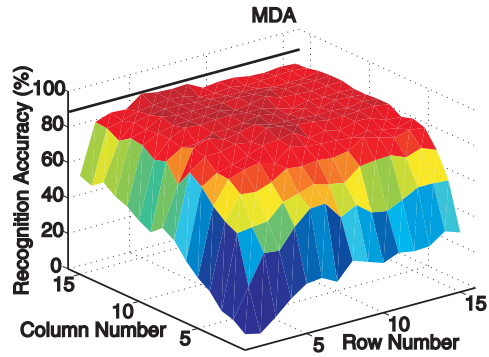,width=7cm}}
  \vspace{0.0cm}
  \centerline{(e) MDA original view}\medskip
\end{minipage}
\hfill
\begin{minipage}[b]{0.48\linewidth}
  \centering
  \centerline{\epsfig{figure=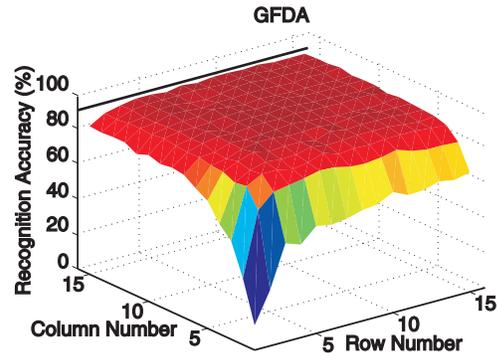,width=7cm}}
  \vspace{0.0cm}
  \centerline{(g) GDA original view}\medskip
\end{minipage}
\hfill
\caption{(Color online) Recognition accuracies (\%) of (2D)$^2$PCA, (2D)$^2$LDA, MDA and GDA versus numbers of features along the row and column directions respectively on ORL database. In GDA, we set the HOSVD threshold (Equation (\ref{equa:ThresholdDimReduction})) 0.98.}
\label{fig:ORLCompare_2D}
\end{figure*}

In order to fairly evaluate the effectiveness of our GDA, we average the recognition accuracies by multiple iterations. Table \ref{tab:AccuracyORL} shows the average face recognition accuracies of all the algorithms in our experiments. The comparative results show our GDA outperforms the other four methods on the three sets of experiment, especially in the cases with a small number of training samples. Fig.~\ref{fig:ORLCompare_2D} demonstrates the accuracies vs. the dimensionality of the four methods on ORL.

\begin{figure}[ht]
\centering
    \begin{minipage}[t]{0.9\linewidth}
        \centering
        \includegraphics[width=1\linewidth]{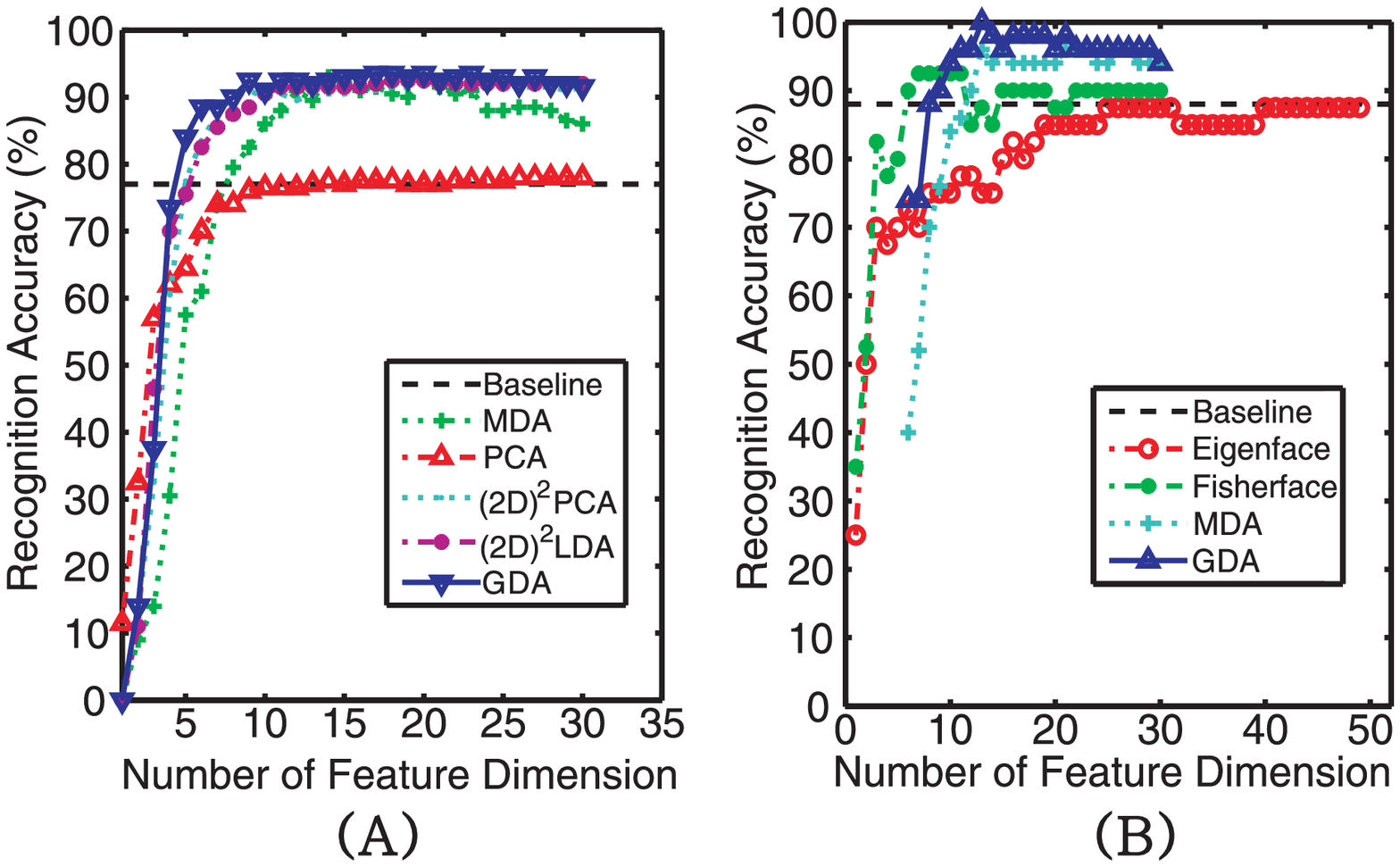}
    \end{minipage}
\caption{Comparisons of recognition accuracies on between MDA, PCA, (2D)$^2$PCA, (2D)$^2$LDA and GDA on ORL database.}
\label{fig:CompareORLline}
\end{figure}

Fig.~\ref{fig:CompareORLline} (A) shows the recognition accuracies of (2D)$^2$PCA, (2D)$^2$LDA, MDA and GDA versus numbers of features along the row and column directions respectively on ORL database, here the training number and test number are both 5.

\begin{table}[ht]
\begin{center}
\caption{ Recognition accuracy (\%) comparison of our proposed GDA with other methods on ORL database }
\label{tab:AccuracyORL}
\begin{tabular}{ c c c c }
\hline\hline

 & Train5/Test5 & Train4/Test6 & Train3/Test7\\
\hline
 PCA          &   90.85        &  87.92  &  83.25   \\
 (2D)$^2$PCA  &   94.70        &  92.58  &  90.36   \\
 (2D)$^2$LDA  &   94.80        &  93.46  &  89.68   \\
 MDA          &   96.50        &  93.42  &  83.30   \\
 GDA         &   \textbf{97.10}        &  \textbf{95.75}  &  \textbf{92.82}   \\
\hline\hline
\end{tabular}
\end{center}
\end{table}

\begin{table*}
\begin{center}
\caption{ Recognition accuracy (\%) comparison of our proposed GDA with other methods on Weizmann database. 
In HOPCA, the first part of GDA, we set threshold $\theta = 0.98$. (CR stands for compression ratio.)}
\label{tab:AccuracyWeizmann}
\begin{tabular}{ c c c c c }
\hline\hline

              & Accuracy  & Dimension & Compression Ratio & Running Time (s) \\
\hline
 Eignface     &   84.4    &   11 & 0.1226      & 0.1198  \\
 Fisherface   &   95.6    &   8 & 0.8894      & 0.1129  \\
 MDA          &   98.89   &  $7\times7\times3$ & 0.0056 & 465.34  \\
 GDA         &   \textbf{98.89}   &  \textbf{6$\times$3$\times$3} &\textbf{0.0028} & \textbf{2.67} \\
\hline\hline
\end{tabular}
\end{center}
\end{table*}

\subsection{Results on Weizmann Database}
\label{ssec:WeizmannExp}
In this subsection, we choose a higher order dataset, Weizmann database\cite{ActionsAsSpaceTimeShapes_iccv05}, to check the quality of our GDA in action recognition.

To compare GDA with other methods fairly, we compute the recognition accuracy using the leave-one-out method. Each time, we first leave out all the sequences pertaining to one person. Then we train using all the remaining sequences (80 sequences), and we use the 10 actions of the omitted person as test actions. We average the results from all the persons.

Table \ref{tab:AccuracyWeizmann} shows the recognition accuracies, where dimension, compression ratio, running time and iteration are acquired when the best accuracy achieves. We can see from the comparative results that our proposed GDA performs the best among all the algorithms.
Even though Eigenface and Fisherface use less running time, their recognition accuracies are far less than that of MDA and GDA. However, MDA directly deal with the high-dimensional data, so the running time becomes much longer. Moreover, the compression ratio acquired in MDA also suffers from the high dimensionality, therefore it achieves no better results than that of GDA.

\section{Conclusion}
\label{sec:Conclusion}
In this paper, we develop Generalized Discriminant Analysis (GDA). GDA provides a more natural representation for images, videos and other high order data, avoiding any models. By analysis, we show GDA enables us to avert the curse of dimensionality and it preserves the spatial and spatial-temporal relationship of the data. Through experiments, we see that GDA can alleviate the small sample size problem and shows high efficiency and effectiveness of computation.

\subsection*{Acknowledgements}
This work is supported by by 973 Program (Project No.2010CB327905) and Natural Science Foundations of China (No.61071218).

\bibliographystyle{ieee}
\bibliography{bib}

\end{document}